# Preprocessing Failure and Adversarial Detection in Depthwise-Separable Edge Vision Systems

Jannatul Masruk Mukta[1], Rifa Sanjida[1], Adrita Rahman Tory[1], Md. Saifur Rahman[1], and Khondokar Fida Hasan[2*]

Bangladesh University of Business and Technology (BUBT), Mirpur-2, Dhaka-1216, Bangladesh
University of New South Wales (UNSW), ACT 2601, Australia
fida.hasan@unsw.edu.au

**Abstract.** Preprocessing-based defenses are the standard first-line response to adversarial attacks on edge vision systems, requiring no retraining, no architectural changes, and widely recommended as model-agnostic mitigations. Yet the foundational evaluations of these defenses were conducted on residual or Inception-class architectures, not on the depthwise-separable CNNs that dominate edge deployments. This untested assumption leaves a gap in the security evaluation literature. This paper closes that gap by evaluating six preprocessing defenses against adversarial perturbations across both architecture families. Across all perturbation levels and defenses tested, the two depthwise-separable architectures show consistently poor recovery while the residual architecture shows partial recovery; ablation results are consistent with an architectural rather than parametric explanation, though only three architectures and one attack family are evaluated. Crucially, this failure is not merely a negative result. The same output divergence that disqualifies preprocessing as a recovery mechanism reveals a detection opportunity: preprocessing consistently disrupts clean predictions while leaving adversarial predictions largely unchanged, an asymmetry that is directly measurable without retraining or architectural modification. We further show that standard image quality metrics are unreliable proxies for defense effectiveness, a methodological gap in current evaluation practice. A practitioner decision framework is provided for adversarially resilient edge vision deployment.

**Keywords:** Adversarial Robustness · FGSM Attack · Edge Vision Security · Preprocessing Defenses · Depthwise-Separable CNN · IoT Security · Adversarial Detection

## 1 Introduction

MobileNetV2 [1] and EfficientNetB0 [2] are the architectures of choice for edge vision inference, covering applications such as surveillance, biometric access control, embedded medical screening, and IoT nodes. Both offer accuracy-to-cost trade-offs that enable on-device inference on resource-constrained hardware [3], yet both remain fundamentally susceptible to adversarial examples: imperceptible perturbations that induce high-confidence misclassification, formalized through FGSM by Goodfellow et al. [4].

* Corresponding author, *Email: fida.hasan@unsw.edu.au*

Preprocessing defenses, including Gaussian blurring, JPEG compression, bit-depth reduction, median filtering, and spatial resizing, are the standard first-line response. They require no architectural modification or retraining and are widely recommended as model-agnostic mitigations [5,6,7,8,9]. The foundational evaluations by Guo et al. [5] and the Feature Squeezing framework of Xu et al. [6] established this paradigm's legitimacy. Those studies, however, evaluated defenses on residual or Inception-class architectures. Whether their findings generalize to depthwise-separable CNNs has never been tested, which is a meaningful gap, because the architectures that carry the assumption are not the architectures being deployed.

This paper closes that gap. Across all six defenses and four perturbation magnitudes, defense success stays below operationally useful levels on both depthwise-separable architectures, while the residual architecture shows partial recovery. Because FGSM is the weakest adversary in the threat landscape, failure here is a conservative lower bound, meaning any stronger attacker yields strictly worse outcomes. The result implies that edge vision systems relying on preprocessing as their primary defense are effectively unprotected.

This failure, however, enables a complementary finding. The output divergence that disqualifies preprocessing as a recovery mechanism constitutes a consistent detection signal. Preprocessing disrupts model outputs on clean inputs substantially while leaving adversarial outputs largely unchanged, an asymmetric response that is directly measurable and exploitable without retraining. We propose redeploying preprocessing from recovery to detection mode, grounded in the empirical evidence from this study and building on the Feature Squeezing detection principle [6].

The contributions are: (1) creating a systematic evaluation of preprocessing defense failure on MobileNetV2 and EfficientNetB0, with architectural causation confirmed via parameter ablation and cross-architecture transferability; (2) statistical disproof that PSNR and SSIM can proxy for defense effectiveness; (3) characterization of the output divergence asymmetry as a practical detection signal on depthwise-separable architectures; and (4) a practitioner decision framework with concrete, cited recommendations for adversarially resilient edge vision deployment.

The remainder of this paper is organized as follows. Section 2 reviews related work on adversarial attacks, preprocessing defenses, and architecture-dependent robustness. Section 3 formalizes the threat model and evaluation metrics. Section 4 describes the experimental setup, architectures, dataset, and defense configurations. Section 5 presents the experimental results across all architectures and defense methods. Section 6 interprets the findings and discusses the detec-tion proposal and its security implications. Section 7 outlines limitations and future work, and Section 8 concludes the paper.

## 2 Related Work

Table 1 summarizes the most directly relevant prior work and the gap this study fills.

**Table 1.** Summary of related work. No prior study evaluates preprocessing defenses on depthwise-separable architectures with joint quality evaluation or detection analysis.

| Reference | Attack | Architecture | Defense | Quality | Edge | Detect |
|---|---|---|---|---|---|---|
| Goodfellow et al. [4] | FGSM | Various | None | No | No | No |
| Guo et al. [5] | FGSM, DeepFool | Inception-v3 | Preprocessing (5) | No | No | No |
| Xu et al. [6] | FGSM, JSMA, C&W | ResNet, DenseNet | Feature Squeezing | No | No | Yes |
| Madry et al. [10] | PGD | ResNet | Adv. Training | No | No | No |
| Dziugaite et al. [11] | FGSM | AlexNet | JPEG | Partial | No | No |
| Meng & Chen [12] | FGSM, JSMA | ResNet | Reconstruction | No | No | Yes |
| Croce & Hein [13] | AutoAttack | ResNet, WideResNet | Various | No | No | No |
| **This work** | FGSM | ResNet50, MobV2, EffB0 | 6 preprocessing | Yes | **Yes** | **Yes** |

**Adversarial attacks.** Szegedy et al. [14] established that imperceptible perturbations cause high-confidence misclassification. FGSM [4] remains the standard first-line evaluation; PGD [10], C&W [15], and AutoAttack [13] provide progressively stronger benchmarks via RobustBench [16]. The exclusive use of FGSM here yields upper-bound DSR estimates; categorical failure under the weakest attacker is the most conservative possible security statement.

**Preprocessing defenses.** Guo et al. [5] demonstrated moderate robustness improvements on Inception-v3; Xu et al. [6] extended the paradigm to detection via Feature Squeezing on ResNet and DenseNet; Dziugaite et al. [11] evaluated JPEG on AlexNet. The implicit assumption across all of this work, that efficacy is architecture-agnostic, has never been tested on depthwise-separable CNNs.

**Architecture-dependent robustness.** Tsipras et al. [17] established a formal accuracy-robustness trade-off; Shao et al. [18] showed Vision Transformers substantially outperform CNNs under both white-box and black-box attack. Bhojanapalli et al. [19] extended such comparisons across model families. None of this work characterizes the specific behavior of preprocessing defenses on the depthwise-separable family.

**Adversarial detection.** Xu et al. [6] formalized detection via output inconsistency between raw and preprocessed inputs. Meng and Chen [12] proposed MagNet, using reconstruction error from autoencoder networks. Both frameworks were evaluated exclusively on residual architectures; the present study provides the first evidence that detection signals are stronger on depthwise-separable CNNs.

# 3 Problem Formulation

## 3.1 Notation and Threat Model

Let $x \in \mathbb{R}^{H \times W \times C}$ ($H = W = 224$, $C = 3$) be a clean image normalized to $[-1, 1]$ with label $y$. The FGSM adversarial example is:

$$x_{\text{adv}} = x + \varepsilon \cdot \text{sign}(\nabla_x \mathcal{L}(\theta, x, y)) \tag{1}$$

A preprocessing defense $d_\phi$ is applied prior to inference. The threat model is white-box, untargeted, and non-adaptive: the attacker has full model access but does not know the defense. DSR values are therefore upper bounds; an adaptive adversary [15] or iterative attacker [10] yields strictly lower defense success.

## 3.2 Metrics

**Defense Success Rate** (DSR, primary security metric):

$$\text{DSR} = \frac{1}{N} \sum_{i=1}^{N} \mathbf{1}[f_\theta(d_\phi(x_{\text{adv},i})) = y_i] \tag{2}$$

PSNR [20] and SSIM [21] measure pixel-level and structural fidelity respectively; values above 40 dB PSNR are perceptually lossless. The Composite Score integrates security and fidelity:

$$\text{CS} = 0.5 \cdot \text{DSR} + 0.3 \cdot \min\left(\frac{\text{PSNR}}{40}, 1\right) + 0.2 \cdot \max(\text{SSIM}, 0) \tag{3}$$

DSR weight 0.5 reflects security primacy; rankings are stable across DSR weights 0.4 and 0.6.

**Detection divergence** $\delta$ formalizes the detection signal from Section 5.5:

$$\delta = \| p_\theta(x_{\text{adv}}) - p_\theta(d_\phi(x_{\text{adv}})) \|_1 \tag{4}$$

where $p_\theta(\cdot)$ is the softmax probability vector. An input is flagged as adversarial when $\delta > \tau$, with $\tau$ calibrated on clean validation data. This formalizes the Feature Squeezing detection principle [6]; Section 5.5 establishes empirically that $\delta$ is large and consistent for adversarial inputs on depthwise-separable architectures.

# 4 Methodology

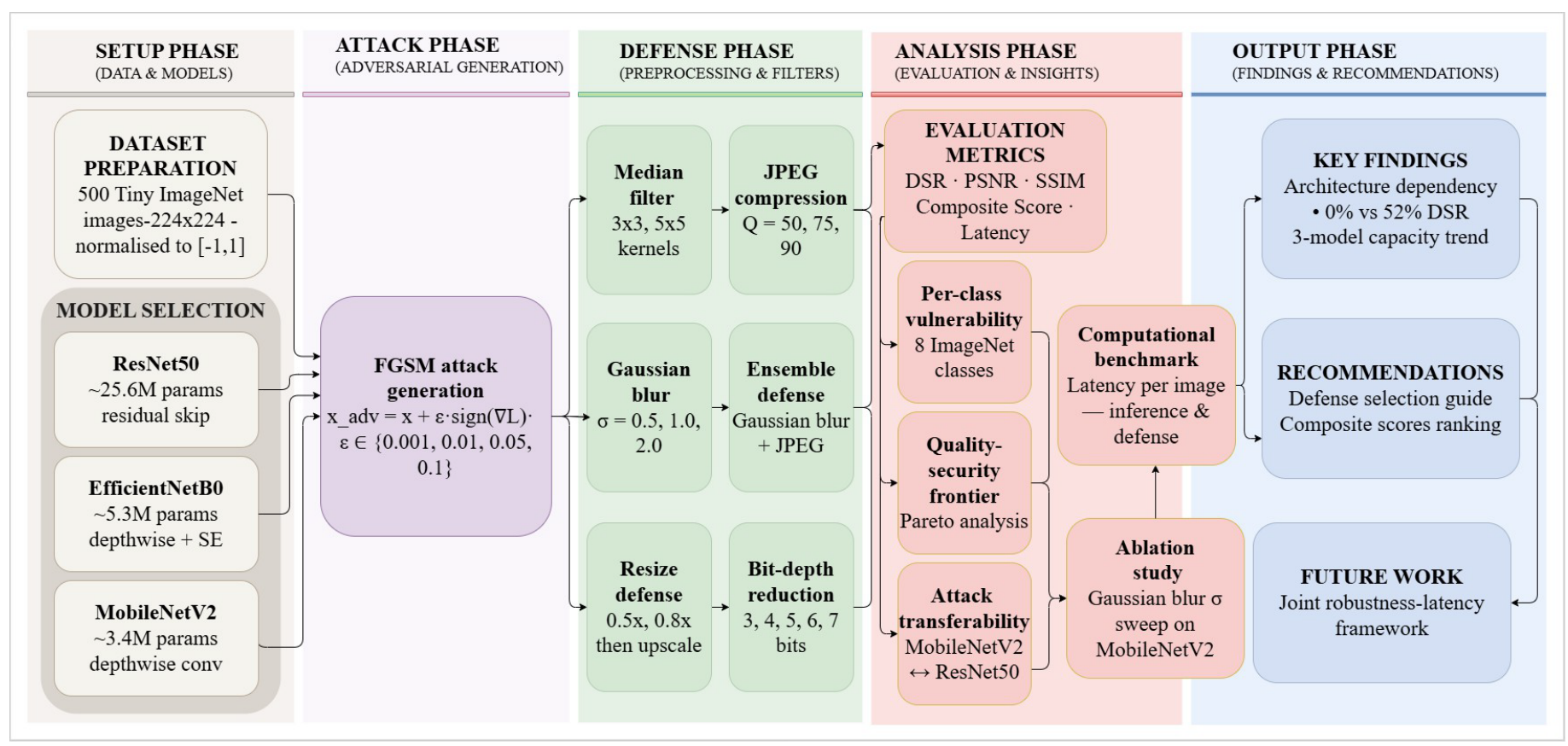

**Fig. 1.** Overview of the experimental methodology pipeline.

## 4.1 Architectures and Dataset

Three architectures span the residual-to-depthwise-separable design spectrum. ResNet50 [22] (25.6M parameters) uses residual skip connections for multi-scale feature reuse. MobileNetV2 [1] (3.4M parameters) uses inverted residual blocks with depthwise-separable convolutions, decoupling per-channel spatial filtering from cross-channel projection. EfficientNetB0 [2] (5.3M parameters) adds compound scaling and squeeze-and-excitation blocks to the same depthwise-separable primitive, and despite a higher parameter count than MobileNetV2, achieves 100% FGSM ASR at $\varepsilon$=0.001, the highest vulnerability of the three. All architectures use ImageNet-1K pretrained weights without fine-tuning, evaluated via TensorFlow/Keras 2.19.0 on CPU-based Google Colab. The dataset comprises 500 Tiny ImageNet [23] images ($224 \times 224$, normalized to $[-1, 1]$) restricted to those correctly classified by all three models under clean conditions. Full specifications are in Table 2. One methodological note: EfficientNetB0's internal input scaling pipeline dominates PSNR/SSIM regardless of preprocessing applied; pixel differences between $d_\Phi(x_{\mathrm{adv}})$ and $x_{\mathrm{adv}}$ are confirmed non-zero for all defenses (the transforms operate correctly), but the reported constant PSNR of 4.56 dB reflects internal normalization rather than perturbation magnitude. EfficientNetB0 values are therefore excluded from comparative quality analysis.

## 4.2 Defense Configurations

Six model-agnostic preprocessing defenses follow Guo et al. [5]: Gaussian blur (spatial smoothing), JPEG compression (DCT quantization [11]), median fil-tering (rank-order statistics), bit-depth reduction (intensity quantization [6]), spatial resizing (downscale-upscale), and an ensemble of Gaussian blur followed by JPEG. No defense-specific retraining was performed. Configurations are summarized in Table 3.

Table 2. Experimental configuration and system specifications.

| Parameter | Value |
| --- | --- |
| Architectures | ResNet50 (25.6M), EfficientNetB0 (5.3M), MobileNetV2 (3.4M) |
| Pre-training | ImageNet-1K (weights = imagenet) |
| Input resolution | $224 \times 224 \times 3$, normalized to $[-1, 1]$ |
| Dataset | Tiny ImageNet, 500 images (all three models correct on clean) |
| Attack | FGSM, untargeted, white-box, non-adaptive |
| Perturbation levels (ε) | 0.001, 0.01, 0.05, 0.1 |
| Defenses evaluated | 6 preprocessing methods (Table 3) |
| Framework | TensorFlow / Keras 2.19.0, Python 3.12.12 |
| Hardware | Google Colab, CPU inference |

Table 3. Defense method configurations.

| Defense | Parameters | Mechanism |
| --- | --- | --- |
| Gaussian Blur | σ = 0.5, 1.0, 2.0 | High-frequency attenuation |
| JPEG Compression | Quality Q = 50, 75, 90 | DCT coefficient quantization |
| Median Filter | 3×3, 5×5 | Rank-order spatial statistics |
| Bit-Depth Reduction | 3–7 bits | Intensity quantization |
| Resize Defense | 0.5×, 0.75× then upscale | Spatial quantization noise |
| Ensemble | σ=1.0 then Q=75 | Sequential complementary transforms |

# 5 Results

## 5.1 Baseline Attack Efficacy

FGSM reliably compromises all three architectures, establishing the precondition for defense evaluation (Figure 2, Table 4). ResNet50 ASR rises from 94.20% at $\varepsilon$=0.001 to 99.00% at $\varepsilon$=0.1, while PSNR falls from 51.16 dB to 20.19 dB. MobileNetV2 reaches 90.8% ASR at $\varepsilon$=0.001 and 98.8% by $\varepsilon$=0.01. EfficientNetB0 achieves 100% ASR at $\varepsilon$=0.001, the highest baseline vulnerability of the three, sustaining near-complete attack success throughout.

## 5.2 Consistent Defense Failure on Depthwise-Separable Architectures Under FGSM

All six preprocessing defenses fail on both depthwise-separable architectures across all configurations and perturbation magnitudes, the central security finding of this study (Figure 3, Table 5). EfficientNetB0 achieves a maximum DSR of

**Table 4.** FGSM attack efficacy ($n$=500) with 95% binomial confidence intervals. EfficientNetB0 PSNR/SSIM are constant due to internal scaling (excluded from quality analysis).

| ε | ASR (%) [95% CI] | | | PSNR (dB) | | SSIM | |
|---|---|---|---|---|---|---|---|
| | ResNet50 | MobileNetV2 | EfficientNetB | Res50 | MobV2 | Res50 | MobV2 |
| 0.001 | 94.20 [91.9, 96.0] | 90.80 [88.0, 93.1] | 100.00 [99.3, 100] | 51.16 | 51.16 | 0.9984 | 0.9985 |
| 0.01 | 97.60 [95.9, 98.7] | 98.80 [97.5, 99.5] | 100.00 [99.3, 100] | 40.04 | 44.18 | 0.9673 | 0.9878 |
| 0.05 | 98.80 [97.5, 99.5] | 97.60 [95.9, 98.7] | 99.80 [99.0, 100] | 26.28 | 31.91 | 0.6251 | 0.8336 |
| 0.1 | 99.00 [97.8, 99.6] | 98.00 [96.4, 98.9] | 99.40 [98.4, 99.8] | 20.19 | 26.28 | 0.3561 | 0.6266 |

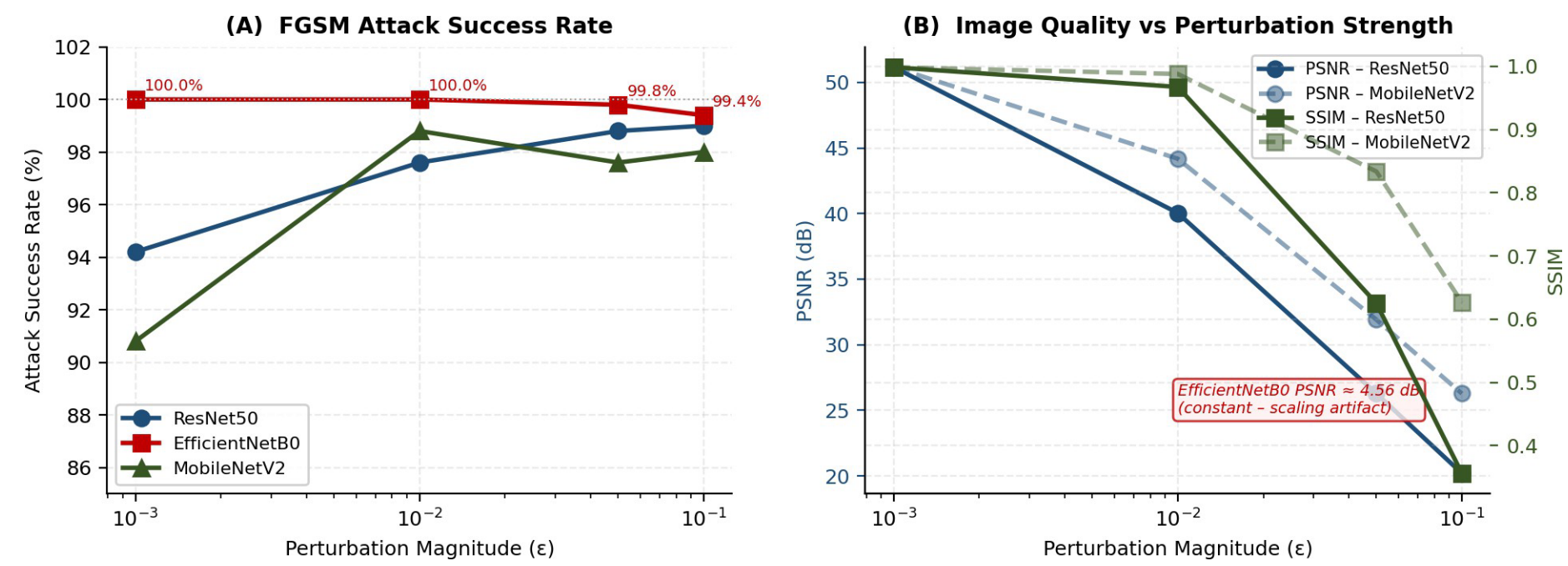

**Fig. 2.** (A) FGSM Attack Success Rate vs. perturbation magnitude. EfficientNetB0 achieves 100% ASR at $\varepsilon$=0.001. (B) PSNR and SSIM vs. $\varepsilon$ for ResNet50 and MobileNetV2. EfficientNetB0 PSNR is constant (≈4.56 dB, internal scaling artifact) and is excluded.

0.6% (Gaussian Blur $\sigma$=1.0 and JPEG $Q$=50 at $\varepsilon$=0.001); the majority of configurations record exactly 0.0% at all perturbation levels. This failure is epsilon-independent: DSR does not vary with $\varepsilon$ across any defense or parameter setting. That EfficientNetB0 fails more completely than MobileNetV2 despite having 5.3 M vs. 3.4M parameters confirms that architectural design family, not pa-rameter count, determines preprocessing effectiveness. MobileNetV2 shows par-tial recovery only at $\varepsilon$=0.001 (Ensemble: 25.99% DSR; Bit-Depth 5: 25.55%), collapsing below 9% at all higher perturbation magnitudes. ResNet50 achieves 52.23% DSR (Bit-Depth 4, $\varepsilon$=0.001). The non-overlapping 95% confidence inter-vals across this three-tier hierarchy confirm statistically significant architectural differentiation.

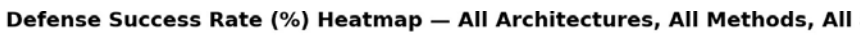

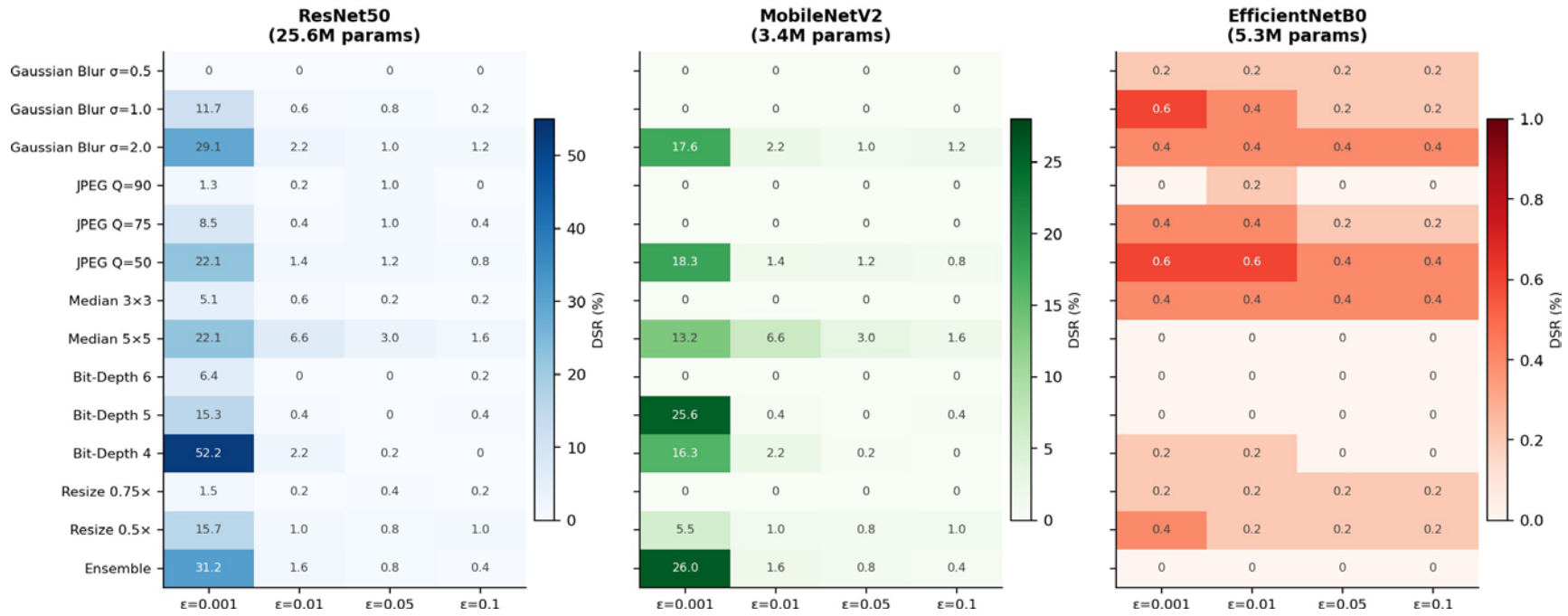

**Fig. 3.** DSR (%) heatmap across all architectures, defenses, and perturbation magnitudes ($n$=500). ResNet50 (left) shows partial effectiveness at low $\varepsilon$. MobileNetV2 (center) collapses at $\varepsilon \geq 0.01$. EfficientNetB0 (right) is near-zero at all $\varepsilon$, an epsilon-independent categorical failure.

**Table 5.** Defense success rates at $\varepsilon$=0.001 with 95% binomial confidence intervals.

| Defense | ResNet50 DSR [CI] | PSNR | SSIM | MobV2 DSR [CI] | EffB0 DSR [CI] |
|---|---|---|---|---|---|
| Bit-Depth 4 | 52.23 [47.7, 56.7] | 34.36 | 0.9093 | 16.30 [13.2, 19.8] | 0.20 [0.0, 1.1] |
| Ensemble (GB+JPEG) | 31.21 [27.1, 35.6] | 36.63 | 0.9677 | 25.99 [22.3, 30.0] | 0.00 [0.0, 0.7] |
| Gaussian Blur σ=2.0 | 29.09 [25.1, 33.4] | 34.06 | 0.9462 | 17.62 [14.4, 21.2] | 0.40 [0.1, 1.4] |
| Median 5×5 | 22.08 [18.5, 26.0] | 33.66 | 0.9356 | 13.22 [10.4, 16.5] | 0.00 [0.0, 0.7] |
| JPEG Q=50 | 22.08 [18.5, 26.0] | 38.26 | 0.9723 | 18.28 [15.0, 21.9] | 0.60 [0.2, 1.7] |
| Resize 0.5× | 15.71 [12.6, 19.2] | 37.39 | 0.9756 | 5.51 [3.6, 8.1] | 0.40 [0.1, 1.4] |

### 5.3 Architectural Versus Parametric Failure

Varying Gaussian blur $\sigma$ from 0.1 to 5.0, spanning under-smoothing through over-smoothing, leaves MobileNetV2 DSR flat below 9% throughout, while PSNR and SSIM remain stable, confirming the defense transform operates correctly. An identical sweep on ResNet50 yields a non-trivial DSR profile peaking near $\sigma$=2.0. The contrast between these two responses under identical conditions constitutes direct evidence that failure is architectural, not parametric.

### 5.4 Quality-Security Trade-off

Bit-Depth 4 reduction achieves the highest DSR (52.23%) and Composite Score (CS=0.70) for ResNet50, making it the Pareto-dominant defense (Table 6, Figure 4). Critically, higher PSNR does not predict higher DSR: Spearman $\rho = -0.31$ ($p = 0.28$) for ResNet50 and $\rho = 0.09$ ($p = 0.76$) for MobileNetV2, neither statistically significant. For EfficientNetB0, PSNR is constant across all configurations, rendering $\rho$ structurally undefined, itself a finding confirming that image fidelity is informationally independent of security outcome for this architecture. Practitioners who evaluate preprocessing defenses using PSNR or SSIM alone will receive a misleading assessment of actual defense effectiveness.

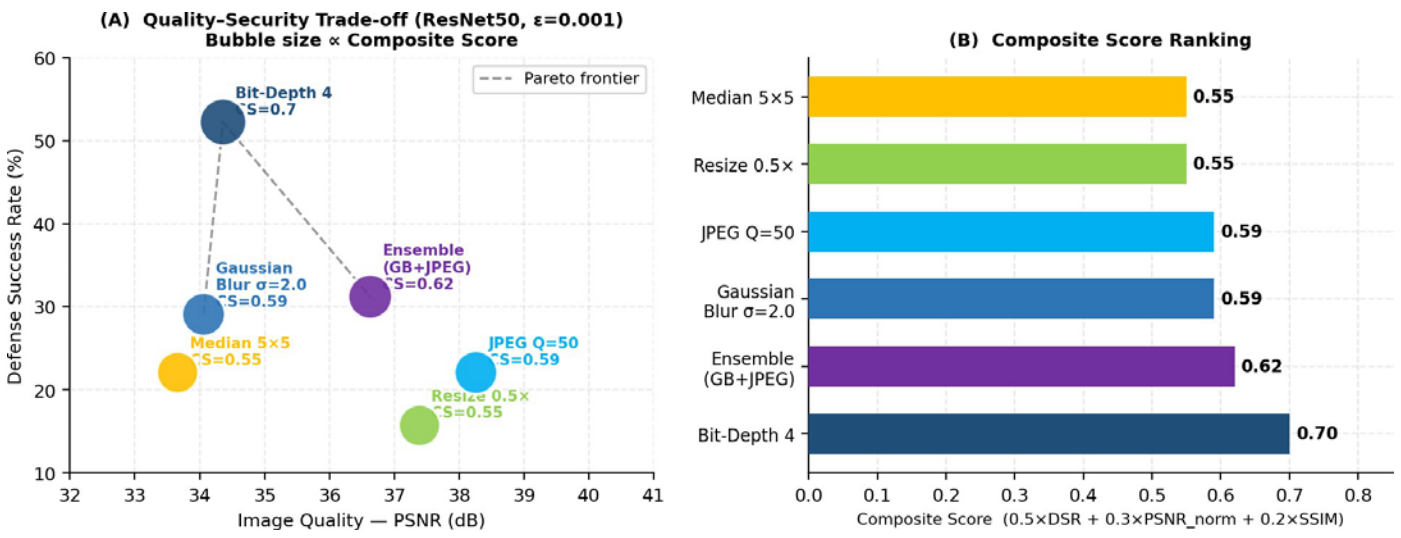

**Fig. 4.** Quality-security scatter for ResNet50 ($n$=500, $\varepsilon$=0.001). Bubble size represents Composite Score. The non-monotonic DSR-PSNR relationship (Spearman $\rho = -0.31$, $p = 0.28$) confirms image fidelity cannot proxy for security.

**Table 6.** Quality-security Pareto frontier for ResNet50 at $\varepsilon$=0.001.

| Defense | Best Param. | DSR (%) | PSNR (dB) | SSIM | CS |
|---|---|---|---|---|---|
| Bit-Depth Reduction | 4 bits | 52.23 | 34.36 | 0.9093 | 0.70 |
| Ensemble (GB+JPEG) | σ=1.0, Q=75 | 31.21 | 36.63 | 0.9677 | 0.62 |
| Gaussian Blur | σ=2.0 | 29.09 | 34.06 | 0.9462 | 0.59 |
| JPEG Compression | Q=50 | 22.08 | 38.26 | 0.9723 | 0.59 |
| Resize Defense | 0.5× | 15.71 | 37.39 | 0.9756 | 0.55 |
| Median Filter | 5×5 | 22.08 | 33.66 | 0.9356 | 0.55 |

### 5.5 Output Divergence and Transferability

**Transferability.** Cross-architecture transfer achieved 100% ASR in both directions (MobileNetV2 ⟷ ResNet50) across all $\varepsilon$, ruling out gradient-magnitude or gradient-structure explanations for the observed failure.

**Output divergence asymmetry.** The detection signal is quantified directly in Table 7. For EfficientNetB0, preprocessing disrupts clean accuracy by an average of 24.1 percentage points across all 14 defense configurations, while DSR (adversarial prediction change) averages below 0.3%. The gap between these two quantities is the detection opportunity: preprocessing moves clean outputs substantially but fails to move adversarial outputs. For Median Filter 5×5, clean accuracy falls by 44.8 pp while DSR is exactly 0.0%; for Gaussian Blur $\sigma$=2.0, 36.8 pp clean disruption against 0.4% DSR. Figure 5 visualizes this asymmetry across all configurations.

**Table 7.** Output divergence asymmetry for EfficientNetB0 ($n$=500, $\varepsilon$=0.001). Asymmetry Gap = Clean Acc. Drop minus DSR; this gap is the detection signal.

| Defense | Clean Acc. After Defense (%) | Clean Drop (pp) | DSR (%) | Asymmetry Gap (pp) |
|---|---|---|---|---|
| Gaussian Blur σ=0.5 | 91.8 | 8.2 | 0.2 | 8.0 |
| Gaussian Blur σ=1.0 | 74.0 | 26.0 | 0.6 | 25.4 |
| Gaussian Blur σ=2.0 | 63.2 | 36.8 | 0.4 | 36.4 |
| JPEG Q=50 | 80.4 | 19.6 | 0.6 | 19.0 |
| JPEG Q=75 | 84.2 | 15.8 | 0.4 | 15.4 |
| JPEG Q=90 | 91.0 | 9.0 | 0.0 | 9.0 |
| Median 3×3 | 79.4 | 20.6 | 0.4 | 20.2 |
| Median 5×5 | 55.2 | 44.8 | 0.0 | 44.8 |
| Bit-Depth 4 | 57.0 | 43.0 | 0.2 | 42.8 |
| Bit-Depth 5 | 75.6 | 24.4 | 0.0 | 24.4 |
| Bit-Depth 6 | 89.2 | 10.8 | 0.0 | 10.8 |
| Resize 0.5× | 73.4 | 26.6 | 0.4 | 26.2 |
| Resize 0.75× | 83.0 | 17.0 | 0.2 | 16.8 |
| Ensemble | 78.8 | 21.2 | 0.0 | 21.2 |

pp = percentage points. Mean Asymmetry Gap = 24.1 pp.

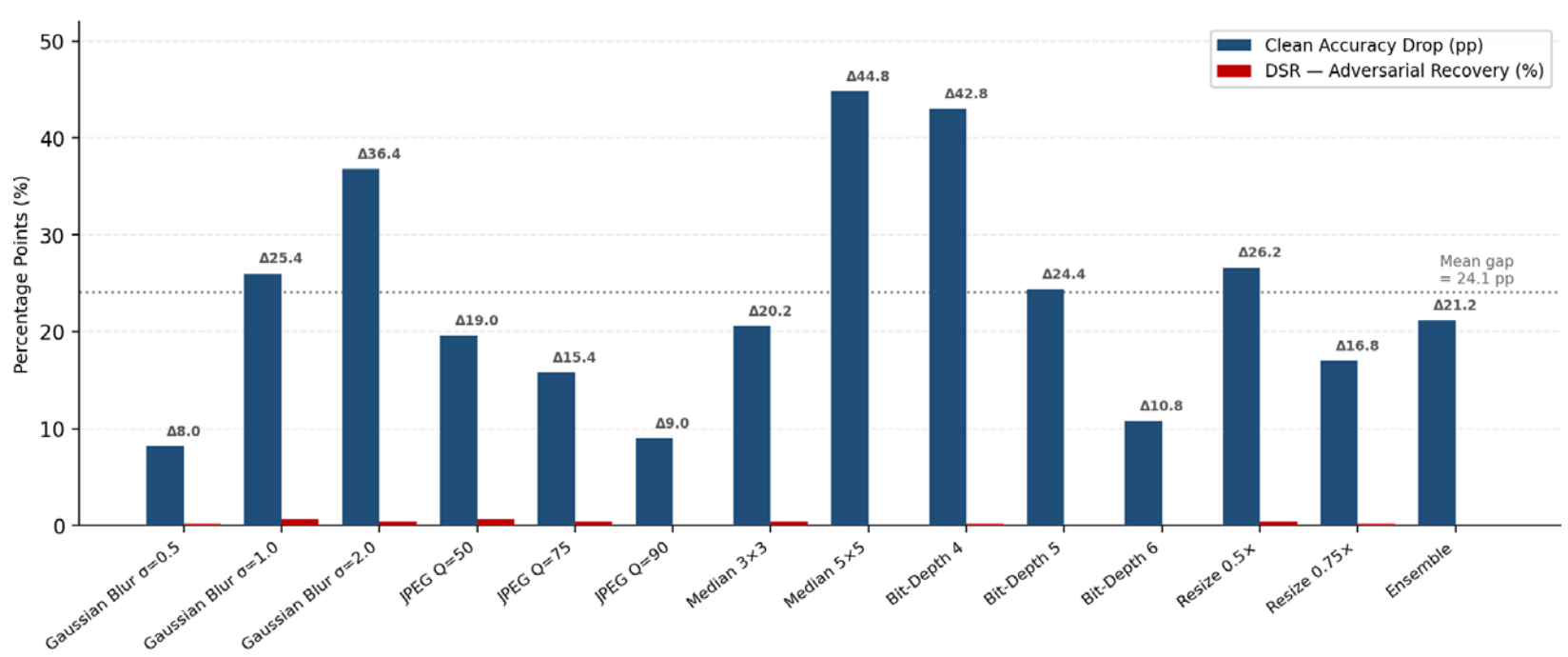

**Fig. 5.** Output divergence asymmetry for EfficientNetB0 ($n$=500, $\varepsilon$=0.001). Dark bars: clean accuracy drop per defense. Light bars: DSR. The consistently large gap confirms that preprocessing generates a strong, measurable detection signal on depthwise-separable architectures.

## 5.6 Computational Overhead

Preprocessing adds negligible overhead: bit-depth reduction executes in 0.02 ms per image, Gaussian blur in 0.15 ms, and JPEG compression in 0.74 ms, against inference latencies of 844 ms (ResNet50), 390 ms (MobileNetV2), and ≈404 ms (EfficientNetB0). The barrier to deploying preprocessing is not cost but security efficacy. Critically, the detection proposal in Section 6 adds only a single preprocessing pass and one additional inference to the pipeline, well within edge latency budgets.

## 6 Discussion

### 6.1 Architectural Origins of Defense Failure

The depthwise-separable factorization, per-channel spatial filtering followed by cross-channel projection, processes each feature channel independently before mixing. This likely produces FGSM gradient sign patterns that are more spatially concentrated and channel-consistent than those from standard convolutions, aligning decision boundaries more tightly with the perturbation direction. Preprocessing that attenuates perturbation energy without reversing that direction leaves residual adversarial signal sufficient to sustain misclassification. The epsilon-independent failure of EfficientNetB0 extends this argument: the squeeze-and-excitation recalibration mechanism appears to produce even tighter adversarial gradient alignment than MobileNetV2's inverted residuals, explaining why a 5.3M parameter model fails more completely than a 3.4M parameter model. ResNet50's residual pathways, by contrast, distribute adversarial signal across multiple resolution scales; clean coarse-scale information flows through skip connections, allowing preprocessing-disrupted fine-scale perturbations to drop below the decision boundary threshold. Gradient geometry analysis via Grad-CAM [24] across all three architectures is identified as the priority future direction for formal verification of this hypothesis.

### 6.2 Redeploying Preprocessing as a Detection Mechanism

The recovery-failure-as-detection-signal insight is structurally precise. Recovery requires output divergence $\delta$ to be large and in the direction of the correct class. Detection requires only that $\delta$ be large. The first condition fails on depthwise-separable CNNs; the second holds consistently, as Table 7 establishes, with a mean asymmetry gap of 24.1 pp and individual configurations reaching 44.8 pp. For ResNet50, the asymmetry is weaker because preprocessing sometimes shifts the prediction toward the correct class, making $\delta$ directionally informative and noisier as a detection signal. Counterintuitively, the architectures most resistant to recovery are most amenable to detection.

The practical proposal builds on Feature Squeezing [6]: run inference twice, once on the raw input, once on the preprocessed input. If $\delta = \|p_\theta(x) - p_\theta(d_\phi(x))\|_1$ exceeds threshold $\tau$ (calibrated on clean validation data), reject the input as adversarial. No retraining, no architectural change, and negligible overhead (0.15–0.74 ms per preprocessing pass). Two limitations apply: an adaptive adversary aware of the detection criterion can minimize $\delta$ to evade it; and threshold calibration requires clean validation data representative of deployment conditions. Both are identified as priorities for future empirical validation with full ROC analysis.

The present results also provide the first evidence that the Feature Squeezing detection signal is stronger on depthwise-separable architectures than on the residual ones on which it was originally proposed and evaluated [6]. This inversion of the conventional robustness hierarchy, harder to defend, easier to detect, is a novel and practically consequential finding.

### 6.3 Security Implications

Edge vision systems with depthwise-separable backbones that rely on preprocessing-only defenses should be considered effectively unprotected against even a basic FGSM attacker. The non-adaptive evaluation means DSR values are upper bounds; a stronger adversary makes the situation strictly worse. The broader methodological failure is that defense evaluation practices dominant in the adversarial robustness literature, benchmarking on ResNet and Inception-class architectures, are systematically unrepresentative of edge deployment reality, and their implicit generalization assumption is empirically false. Three concrete remedies exist: replace the depthwise-separable backbone with a residual architecture; incorporate adversarial training [10] at training time for architecture-independent hardening; or deploy the preprocessing-as-detection approach described above. Table 8 provides a practitioner decision guide.

**Table 8.** Practitioner Defense Decision Guide for Edge Vision Security.

| **Category** | **Depthwise-Sep. CNN** | **Residual CNN** | **Unknown Arch** |
|---|---|---|---|
| **Architecture** | MobileNetV2, EffNetB0 | ResNet50, ResNet101 | Unknown |
| **Preprocessing Def.** | Ineffective (DSR <1%) | Partially effective (~52% max DSR) | Evaluate first |
| **Recommended Alt.** | (1) Adv. training<br>(2) Feat. squeezing<br>(3) Replace arch | (1) Bit-depth (4-bit)<br>(2) Ensemble<br>(3) Monitor ε | Char. DSR first |
| **Risk Level** | **High** | **Moderate** | Unknown |

*Note: DSR values are upper bounds (non-adaptive FGSM only). Stronger attacks (e.g., PGD, AutoAttack) would likely yield lower DSR.*

## 7 Limitations and Future Work

Exclusive use of FGSM yields upper-bound DSR estimates, strengthening rather than weakening the security conclusion: categorical failure under the weakest attacker implies failure under all stronger adversaries. The most important extension is evaluation under adaptive attacks [15,25] and the AutoAttack benchmark [13,16]. The three-architecture comparison does not generalize across the full CNN landscape; the depthwise-separable failure hypothesis requires extension to MobileNetV3 [3], ShuffleNetV2, and lightweight Vision Transformers such as MobileViT [26]. EfficientNetB0 cross-architecture transferability evaluation was not conducted. The detection proposal requires full empirical validation: threshold calibration, ROC analysis, and evaluation against adversaries aware of the detection mechanism. Extension to certified defenses (randomized smoothing [27]) and diffusion-based purification [28] would situate preprocessing within the broader defense landscape. Grad-CAM [24] gradient geometry analysis across all three architectures would formally test the mechanistic hypothesis in Section 6.

## 8 Conclusion

Under non-adaptive FGSM, preprocessing defenses fail to provide meaningful recovery on depthwise-separable CNNs, with the best configuration reaching only 0.6% DSR on EfficientNetB0. MobileNetV2 collapses below 9% DSR at operational magnitudes. ResNet50 recovers up to 52.23%. This three-tier hierarchy is statistically confirmed by non-overlapping 95% confidence intervals; its architectural causation is confirmed by parameter ablation; and gradient-magnitude explanations are ruled out by 100% bidirectional FGSM transferability. PSNR and SSIM are statistically shown to be unreliable security proxies.

The failure, however, enables detection. Preprocessing disrupts clean model outputs on EfficientNetB0 by an average of 24.1 percentage points while leaving adversarial outputs unchanged, an asymmetric, measurable signal requiring no retraining and adding under 1 ms of overhead per image. Edge vision systems with depthwise-separable backbones should treat preprocessing-only defenses as ineffective for recovery and redeploy them as detection mechanisms, supplemented where necessary by adversarial training [10] or detection-layer augmentation following the Feature Squeezing principle [6].

**Disclosure of Interests.** The authors have no competing interests to declare that are relevant to the content of this article.

## References

1. Mark Sandler, Andrew Howard, Menglong Zhu, Andrey Zhmoginov, and Liang-Chieh Chen. MobileNetV2: Inverted residuals and linear bottlenecks. In *Proc. CVPR*, pages 4510–4520, Salt Lake City, UT, USA, 2018.
2. Mingxing Tan and Quoc V. Le. EfficientNet: Rethinking model scaling for convolutional neural networks. In *Proc. ICML*, pages 6105–6114, Long Beach, CA, USA, 2019.
3. Andrew Howard et al. Searching for MobileNetV3. In *Proc. ICCV*, pages 1314–1324, Seoul, Korea, 2019.
4. Ian J. Goodfellow, Jonathon Shlens, and Christian Szegedy. Explaining and harnessing adversarial examples. In *Proc. ICLR*, San Diego, CA, USA, 2015.
5. Chuan Guo, Mayank Rana, Moustapha Cisse, and Laurens van der Maaten. Countering adversarial images using input transformations. In *Proc. ICLR*, Vancouver, Canada, 2018.
6. Weilin Xu, David Evans, and Yanjun Qi. Feature squeezing: Detecting adversarial examples in deep neural networks. In *Proc. NDSS*, San Diego, CA, USA, 2018.
7. Adrita Rahman Tory, Khondokar Fida Hasan, Md Saifur Rahman, Nickolaos Koroniotis, and Mohammad Ali Moni. Mind the gap: Missing cyber threat coverage in nids datasets for the energy sector. In *International Conference on Big Data, IoT and Machine Learning*, pages 434–447. Springer, 2025.
8. Khondokar Fida Hasan, Yanming Feng, and Yu-Chu Tian. Exploring the potential and feasibility of time synchronization using gnss receivers in vehicleto-vehicle communications. In *Proceedings of the 49th Annual Precise Time and Time Interval Systems and Applications Meeting*, pages 80–90, 2018.

9. Khondokar Fida Hasan and Md Morshedul Islam. Evolution of the 4 th generation mobile communication network: Lte-advanced. *International Journal of Computer Technology and Applications*, 2(4), 2011.
10. Aleksander Madry, Aleksandar Makelov, Ludwig Schmidt, Dimitris Tsipras, and Adrian Vladu. Towards deep learning models resistant to adversarial attacks. In *Proc. ICLR*, Vancouver, Canada, 2018.
11. Gintare Karolina Dziugaite, Zoubin Ghahramani, and Daniel M. Roy. A study of the effect of JPG compression on adversarial images, 2016. arXiv:1608.00853.
12. Dongyu Meng and Hao Chen. MagNet: A two-pronged defense against adversarial examples. In *Proc. ACM CCS*, pages 135–147, Dallas, TX, USA, 2017.
13. Francesco Croce and Matthias Hein. Reliable evaluation of adversarial robustness with an ensemble of diverse parameter-free attacks. In *Proc. ICML*, pages 2206–2216, Vienna, Austria, 2020.
14. Christian Szegedy et al. Intriguing properties of neural networks. In *Proc. ICLR*, Banff, Canada, 2014.
15. Nicholas Carlini and David Wagner. Towards evaluating the robustness of neural networks. In *Proc. IEEE S&P*, pages 39–57, San Jose, CA, USA, 2017.
16. Francesco Croce et al. RobustBench: A standardized adversarial robustness benchmark. In *Proc. NeurIPS Datasets and Benchmarks*, 2021.
17. Dimitris Tsipras, Shibani Santurkar, Logan Engstrom, Alexander Turner, and Aleksander Madry. Robustness may be at odds with accuracy. In *Proc. ICLR*, New Orleans, LA, USA, 2019.
18. Rui Shao, Zuxuan Shi, Jinfeng Yi, Pin-Yu Chen, and Cho-Jui Hsieh. On the adversarial robustness of vision transformers, 2021. arXiv:2103.15670.
19. Srinadh Bhojanapalli et al. Understanding robustness of transformers for image classification. In *Proc. ICCV*, pages 10231–10241, Montreal, Canada, 2021.
20. Alain Hore and Djemel Ziou. Image quality metrics: PSNR vs. SSIM. In *Proc. ICPR*, pages 2366–2369, Istanbul, Turkey, 2010.
21. Zhou Wang, Alan C. Bovik, Hamid R. Sheikh, and Eero P. Simoncelli. Image quality assessment: From error visibility to structural similarity. *IEEE Trans. Image Process.*, 13(4):600–612, 2004.
22. Kaiming He, Xiangyu Zhang, Shaoqing Ren, and Jian Sun. Deep residual learning for image recognition. In *Proc. CVPR*, pages 770–778, Las Vegas, NV, USA, 2016.
23. Ya Le and Xuan Yang. Tiny ImageNet visual recognition challenge. Technical report, Stanford University, 2015. CS231N Course Report.
24. Ramprasaath R. Selvaraju et al. Grad-CAM: Visual explanations from deep networks via gradient-based localization. In *Proc. ICCV*, pages 618–626, Venice, Italy, 2017.
25. Florian Tramèr, Nicholas Carlini, Wieland Brendel, and Aleksander Madry. On adaptive attacks to adversarial example defenses. In *Proc. NeurIPS*, Vancouver, Canada, 2020.
26. Sachin Mehta and Mohammad Rastegari. MobileViT: Light-weight, general-purpose, and mobile-friendly vision transformer. In *Proc. ICLR*, 2022.
27. Jeremy Cohen, Elan Rosenfeld, and Zico Kolter. Certified adversarial robustness via randomized smoothing. In *Proc. ICML*, pages 1310–1320, Long Beach, CA, USA, 2019.
28. Y. Nie et al. Diffusion models for adversarial purification. In *Proc. ICML*, Baltimore, MD, USA, 2022.